\documentclass[final,1p,times]{elsarticle}
\usepackage{booktabs}
\usepackage{amssymb}
\usepackage{amsmath}
\usepackage{hyperref}
\usepackage{orcidlink}
\usepackage{url}
\usepackage[numbers]{natbib}
\usepackage{epstopdf}
\usepackage{ragged2e}
\usepackage{algorithm}
\usepackage{algpseudocode}
\usepackage{enumerate}
\usepackage{subfigure}
\usepackage{bm}
\usepackage{amsmath,amssymb,amsfonts}%
\usepackage{amsthm}%
\usepackage{mathrsfs}%
\usepackage{stfloats}
\usepackage{color}
\usepackage{colortbl}
\usepackage{pifont}
\usepackage{multirow}
\usepackage{makecell}
\usepackage{footnote}
\usepackage{siunitx}
\usepackage{pifont}
\usepackage{xurl}
\usepackage{xcolor}

\journal{Engineering Applications of Artificial Intelligence}

\begin{document}

\begin{frontmatter}

%% Title, authors and addresses

%% use the tnoteref command within \title for footnotes;
%% use the tnotetext command for theassociated footnote;
%% use the fnref command within \author or \affiliation for footnotes;
%% use the fntext command for theassociated footnote;
%% use the corref command within \author for corresponding author footnotes;
%% use the cortext command for theassociated footnote;
%% use the ead command for the email address,
%% and the form \ead[url] for the home page:
%% \title{Title\tnoteref{label1}}
%% \tnotetext[label1]{}
%% \author{Name\corref{cor1}\fnref{label2}}
%% \ead{email address}
%% \ead[url]{home page}
%% \fntext[label2]{}
%% \cortext[cor1]{}
%% \affiliation{organization={},
%%             addressline={},
%%             city={},
%%             postcode={},
%%             state={},
%%             country={}}
%% \fntext[label3]{}

\title{Physical Operating-Condition-Guided Vision-Language Matching for Cavitation Diagnosis in Industrial Pipeline Systems}

% use optional labels to link authors explicitly to addresses:
\author[1,3]{Yu Sha \orcidlink{0000-0003-4521-2077}}
\author[1]{Junqi Tao \orcidlink{0000-0002-4406-2284}}
\author[1]{Deyu Wang \orcidlink{0009-0001-8544-019X}}
\author[3,4,5]{Horst Stoecker \orcidlink{0000-0002-3282-3664}}
\author[6]{Domagoj Vnucec \orcidlink{0000-0002-2752-9612}}
\author[6]{Andreas Widl \orcidlink{0009-0009-0664-2024}}
\author[1,2,3]{Kai Zhou \orcidlink{0000-0001-9859-1758}}

\affiliation[1]{
            organization={School of Science and Engineering, The Chinese University of Hong Kong},
            city={Shenzhen},
            postcode={518172},
            country={China}}

\affiliation[2]{
            organization={School of Artificial Intelligence, The Chinese University of Hong Kong},
            city={Shenzhen},
            postcode={518172},
            country={China}}            

\affiliation[3]{
            organization={Frankfurt Institute for Advanced Studies},
            city={Frankfurt am Main},
            postcode={60438},
            country={Germany}}  

\affiliation[4]{
            organization={Institut f{\"u}r Theoretische Physik, Goethe Universit{\"a}t Frankfurt},
            city={Frankfurt am Main},
            postcode={60314},
            country={Germany}}  

\affiliation[5]{
            organization={GSI Helmholtzzentrum f{\"u}r Schwerionenforschung GmbH},
            city={Darmstadt},
            postcode={64291},
            country={Germany}}  

\affiliation[6]{
            organization={SAMSON AG},
            city={Frankfurt am Main},
            postcode={60314},
            country={Germany}}  

%% Abstract
\begin{abstract}
Cavitation diagnosis in industrial pipeline systems is challenging because acoustic responses vary substantially with operating conditions, while similar time-frequency patterns may correspond to different cavitation states. To address this issue, we propose a physical operating-condition-guided vision-language matching (POC-VLM) for cavitation diagnosis. POC-VLM first constructs physical operating-condition-aware semantic descriptions by explicitly incorporating sample-specific operating variables into candidate cavitation texts. A time-frequency-conditioned text adaptation module then dynamically aligns these physical semantics with the observed acoustic response, followed by global-local vision-language matching to capture both holistic and localized discriminative patterns. In parallel, a K-nearest neighbours-based non-parametric physical operating-condition prior estimates the physical plausibility of each candidate state and is fused with the vision-language semantic evidence at the decision level. Experiments on three real-world cavitation datasets show that POC-VLM consistently outperforms representative data-driven, multimodal and domain-specific diagnostic methods. Further analyses verify the complementary roles of vision-language semantic matching and physical operating-condition priors, while demonstrating robust performance under varying sampling rates, noise levels and observation durations.
\end{abstract}

%% Keywords

\end{frontmatter}

%% Add \usepackage{lineno} before \begin{document} and uncomment 
%% following line to enable line numbers
%% \linenumbers

%% main text
%%

%% Use \section commands to start a section
\section{Introduction}
\label{sec: 1_introduction}
Cavitation is a common and potentially destructive phenomenon in industrial fluid transport systems, particularly around control valves, pumps and other throttling components \cite{sha2022regional,sha2022acoustic}. The repeated formation and collapse of vapour cavities can generate strong pressure fluctuations, vibration, noise and material erosion, ultimately affecting equipment reliability, operational efficiency and service life \cite{romagnuolo2024study,kalikatzarakis2023physically}. Therefore, accurate identification of different cavitation intensity states is important for condition monitoring and predictive maintenance of industrial pipeline systems \cite{sha2022multi,wang2025cavitation}. Among the available sensing modalities, acoustic signals provide a non-invasive and highly responsive means of capturing transient cavitation behaviour and have consequently been widely used for intelligent cavitation diagnosis \cite{vcdina2003detection,prasshanth2024fault}.

Recent advances in deep learning have substantially improved the capability of extracting discriminative representations from acoustic signals \cite{sha2022multi,peng2026phoneme,zhang2024cfenet}. Convolutional neural networks and Transformer-based architectures can learn hierarchical patterns from raw signals or their time-frequency representations, while more recent fault diagnosis methods further incorporate task-specific structures or hierarchical knowledge to improve the discrimination of different fault intensities. However, most existing approaches remain predominantly data-driven and learn a direct mapping from observed acoustic patterns to diagnostic labels. In practical pipeline systems, cavitation responses are strongly affected by physical operating conditions such as valve stroke, upstream and downstream pressures, flow rate and temperature \cite{romagnuolo2024study,sun2025cavitation}. Therefore, similar acoustic patterns may have different diagnostic implications under different operating conditions, while the same cavitation state may exhibit substantially different responses as the operating condition changes \cite{zhang2026attention,sun2023intelligent,zhang2023multi}. This dependence makes it difficult for purely data-driven models to learn stable diagnostic boundaries across complex operating regimes \cite{dai2023smart,zhang2025mechanical}.

A natural solution is to jointly exploit acoustic responses and physical operating-condition information. Existing multimodal approaches typically encode physical variables as an additional numerical modality and combine them with signal representations through feature concatenation, attention or latent-space interaction \cite{zhang2024time,zhao2025multimodal,yu2026mlfnet}. Although such strategies can improve diagnostic performance, physical variables are often treated merely as auxiliary features rather than as explicit physical context that constrains the interpretation of acoustic observations \cite{,zhang2025multimodal,lu2023physics,xia2026multi}. In particular, conventional feature-level fusion does not directly clarify how a candidate diagnostic state should be interpreted under the current operating condition or whether the predicted state is physically plausible given the observed operating state. To address these questions, physical information should participate not only in representation learning but also in semantic interpretation and final diagnostic reasoning.

Vision-language models provide a promising perspective for this issue. Contrastive Language-Image Pre-training (CLIP) maps visual and textual inputs into a shared semantic space, allowing classification to be reformulated as matching between an observed input and candidate textual descriptions \cite{radford2021learning,zhou2022conditional}. This paradigm provides an opportunity to move beyond fixed discriminative classifiers and represent diagnostic states through semantically meaningful descriptions. However, conventional CLIP prompts are generally static and encode generic class-level semantics \cite{zhou2022conditional}. Such descriptions are insufficient for condition-dependent industrial diagnosis, where the meaning of an acoustic response is inseparable from its physical operating environment. In addition, directly applying pretrained vision-language representations to acoustic time-frequency images introduces a substantial domain gap between natural image-text pretraining and industrial signal interpretation \cite{kim2024evaluating,qiao2024sound,QAID2026113131}. Therefore, effective adaptation requires both physically grounded textual semantics and explicit interaction between candidate semantics and acoustic time-frequency responses.

To address these challenges, we propose physical operating-condition-guided vision-language matching (POC-VLM) for cavitation diagnosis in industrial pipeline systems. The central idea is to use physical operating conditions at both the semantic and decision levels. First, instead of conventional static class prompts, we construct sample-specific physical semantic descriptions by dynamically incorporating the operating conditions associated with each acoustic sample into the candidate cavitation texts, allowing the semantic representation of each state to adapt to the current physical context. Second, we perform cross-attention-based global and class-specific local visual evidence matching, enabling the candidate semantics to interact with both the overall time-frequency response and discriminative local regions relevant to each cavitation state. Third, the physical operating conditions are used to construct a non-parametric nearest-neighbour prior, which estimates the physical plausibility of each candidate state from the distribution of training operating conditions. Finally, the resulting semantic evidence and physical prior are integrated at the decision level, enabling cavitation diagnosis to jointly consider acoustic-semantic compatibility and physical plausibility. The proposed POC-VLM is evaluated on three real-world industrial cavitation datasets. Extensive experiments demonstrate that POC-VLM consistently outperforms representative diagnostic methods and exhibits strong robustness and generalization under varying operating and sensing conditions.

The contributions of this paper are summarized as follows:
\begin{itemize}
    \item We construct dynamic physical semantic descriptions by incorporating sample-specific operating conditions into candidate cavitation texts, enabling fine-grained interaction between physical-condition-aware semantics and acoustic visual evidence
    \item We develop a vision-language matching mechanism that aligns candidate semantics with global and class-specific local time-frequency evidence, enabling fine-grained interaction between physical-condition-aware semantics and acoustic visual evidence.
    \item We construct a nearest-neighbour-based physical operating-condition prior from the training-condition distribution and integrate it with vision-language semantic evidence to jointly assess acoustic-semantic compatibility and physical plausibility.
    \item We evaluate POC-VLM on three real-world industrial cavitation datasets from SAMSON AG, demonstrating consistent performance advantages, robustness and generalization across varying operating and sensing conditions.
\end{itemize}

The remainder of this paper is organized as follows. Section \ref{sec: 2_Preliminaries} introduces the preliminaries. Section \ref{sec: 3_Method} presents the proposed POC-VLM framework. Section \ref{sec: 4_ExperimentSetup} describes the experimental setup. Section \ref{sec: 5_ExperimentalResultAnalysis} reports the experimental results and analyses. Section \ref{sec: conclusions} concludes the paper and discusses future work. 
\section{Preliminaries}
\label{sec: 2_Preliminaries}
\subsection{Cavitation Intensity Diagnosis}
\label{subsec: Cavitation Intensity Diagnosis}
Cavitation is a fluid-dynamic phenomenon involving the formation, growth and collapse of vapour cavities when the local liquid pressure decreases to near or below the vapour pressure \cite{sha2025review,sha2022multi}. In the throttling region of control valves in industrial pipeline systems, local flow conditions vary with operating conditions, significantly influencing cavitation development and its acoustic response. Therefore, similar acoustic time-frequency patterns may have different diagnostic implications under different operating conditions, while the same cavitation state may exhibit different response characteristics as the operating condition changes. This operating-condition dependence provides the primary motivation for jointly exploiting acoustic response information and physical operating-condition information in cavitation diagnosis.

In this study, cavitation intensity diagnosis is formulated as the discrimination of four states: incipient cavitation, constant cavitation, choked-flow cavitation and non-cavitation, with turbulent-flow and no-flow conditions grouped into the non-cavitation class. Different intensities of cavitation are defined in Appendix. Each acoustic recording corresponds to a stable flow state during the acquisition period, and its waveform reflects the cavitation-related acoustic response under specific operating conditions. Therefore, this study aims to accurately distinguish different cavitation intensity states by combining acoustic time-frequency information with physical operating-condition information, supporting intelligent monitoring, operation and maintenance of industrial pipeline systems.

\subsection{Vision-Language Representation}
\label{subsec: Vision-Language Representation}
Vision-Language Representation (VLR) aims to learn a unified semantic space in which visual and textual information can be represented in a directly comparable form \cite{radford2021learning,jia2021scaling}. By aligning visual content with corresponding textual descriptions, vision-language models (VLMs) can measure the semantic relationship between visual information and language concepts through feature similarity, reformulating classification as a matching problem between a visual input and a set of candidate textual descriptions \cite{zhai2022lit,zhou2022conditional}. VLMs employ a visual encoder ${E_v}( \cdot )$ and a text encoder ${E_t}( \cdot )$ to extract features from the two modalities. Given an image $I$ and a candidate textual description $T_c$, the vision-language matching score can be expressed as:
\begin{equation}
\label{eq: vision-language matching score}
s(I,{T_c}) = {\bar E}_v(I)^{\rm T}{\bar E}_t({T_c}),
\end{equation}
where $\bar{(\cdot)}$ denotes feature normalization. By mapping visual and textual features into a shared embedding space, different candidate semantics can be compared according to their feature similarity. In this work, pretrained VLRs are employed as a semantic foundation for cavitation diagnosis. Specifically, time-frequency representations obtained from acoustic signals via continuous wavelet transform (CWT) are used as visual inputs, while candidate textual descriptions are constructed for different cavitation states and matched with the corresponding time-frequency representations. Building on this formulation, POC-VLM further incorporates physical operating-condition information through time-frequency-conditioned text adaptation and global-local vision-language matching, enabling the textual semantics to be dynamically adjusted according to the current acoustic response and operating state. In this way, POC-VLM enhances the representation and discrimination of cavitation-related acoustic characteristics under varying operating conditions.

\section{Method}
\label{sec: 3_Method}
\subsection{Overview of POC-VLM}
\label{subsec: Overview of POC-VLM}
\begin{figure*}[htbp]
    \centering
    \includegraphics[width=\textwidth,height=75mm]{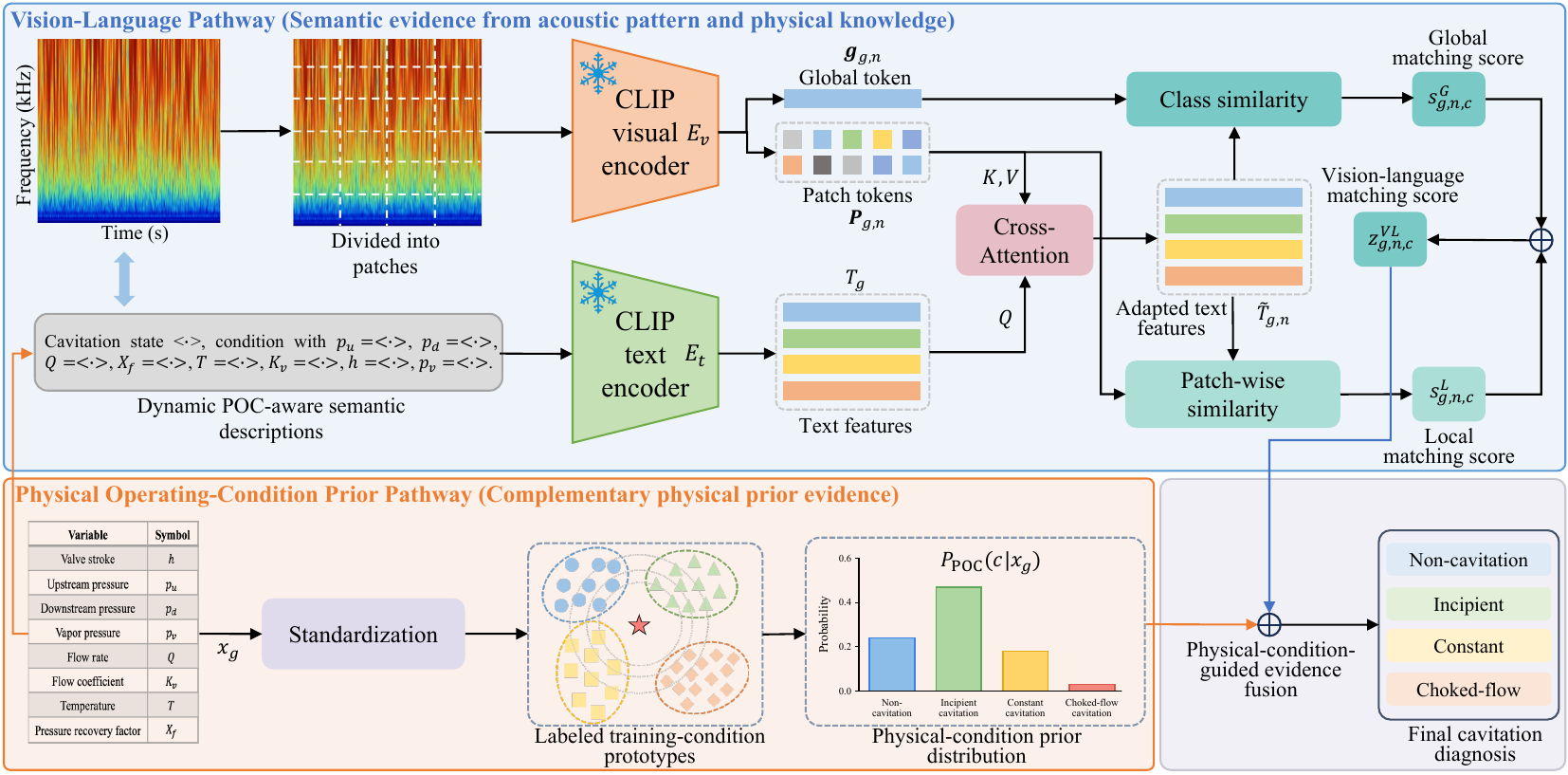}
    \caption{Overall framework of the proposed POC-VLM for cavitation diagnosis. It combines vision-language semantic matching with a physical operating-condition prior through decision-level evidence fusion.}
    \label{fig: Framework}
\end{figure*}

POC-VLM consists of two complementary diagnostic pathways: a vision-language pathway and a physical operating-condition prior pathway, as shown in Figure \ref{fig: Framework}. Let $g$ denote an original acoustic recording, $I_{g,n}$ denote the CWT time-frequency image of its $n$-th window and $\mathbf{x}_g$ denote the physical operating-condition vector shared by all windows from the same recording. For each candidate cavitation class $c$, the vision-language pathway produces a class-wise semantic score through time-frequency-conditioned representation adaptation and global-local matching:
\begin{equation}
z_{g,n,c}^{\text{VL}}=\Phi_{\text{VL}}(I_{g,n},\mathcal{T}(c,\mathbf{x}_g)),
\qquad
c\in\{1,\ldots,C\},
\end{equation}
where $\mathcal{T}(c,\mathbf{x}_g)$ denotes the operating-condition-aware textual description constructed for candidate class $c$ under the corresponding physical operating condition and $\Phi_{\text{VL}}$ represents the complete vision-language matching process. Within this pathway, global image representations characterize the overall time-frequency response, while selected local patch representations provide complementary evidence from discriminative time-frequency regions. The scores obtained for all $C$ candidate classes constitute the vision-language score vector $\mathbf{z}_{g,n}^{\text{VL}}$.

In parallel, the physical operating-condition prior pathway estimates the class prior of the current operating state by comparing $\mathbf{x}_g$ with labeled operating-condition prototypes from the training set:
\begin{equation}
P_{\text{POC}}(c|\mathbf{x}_g)=\Phi_{\text{POC}}(c,\mathbf{x}_g;\{(\mathbf{x}_i,y_i)\}_{i=1}^{M}),
\end{equation}
where $(\mathbf{x}_i,y_i)$ denotes the $i$-th labeled training-condition prototype, $M$ is the total number of prototypes and $\Phi_{\text{POC}}(\cdot)$ represents the non-parametric nearest-neighbour-based prior estimation process. This pathway directly exploits the distribution of training operating conditions to estimate the compatibility of each candidate state with the current operating condition. The semantic evidence from the vision-language pathway and the operating-condition prior are subsequently integrated at the logit level. For each candidate class $c$, the final diagnostic score is defined as:
\begin{equation}
\label{eq: final score}
z_{g,n,c}^{\text{final}}=z_{g,n,c}^{\text{VL}}+\lambda_{\text{POC}}\log[\max(P_{\text{POC}}(c|\mathbf{x}_g),\delta)]
\end{equation}
where $\lambda_{\text{POC}}$ controls the contribution of the physical operating-condition prior and $\delta$ is a small constant introduced for numerical stability. From a probabilistic perspective, Eq. \ref{eq: final score} can be equivalently expressed as:
\begin{equation}
P(c|I_{g,n},\mathbf{x}_g) \propto \exp(z_{g,n,c}^{\text{VL}})P_{\text{POC}}(c|\mathbf{x}_g)^{\lambda_{\text{POC}}}.
\end{equation}

In this formulation, the vision-language pathway provides semantic evidence from the observed acoustic time-frequency response, while the physical operating-condition prior provides complementary evidence from the training operating-condition distribution. Their integration enables the final diagnosis to jointly consider the acoustic response and its associated physical operating state.

\subsection{Physical Operating-Condition-Aware Vision-Language Representation}
\label{subsec: Physical Operating-Condition-Aware Vision-Language Representation}
The physical operating-condition-aware textual description $\mathcal{T}(c,\mathbf{x}_g)$ is constructed by combining each candidate cavitation state $c$ with the physical operating variables associated with the current acoustic recording. Specifically, the textual description is formulated as: \textit{Cavitation state $<\cdot>$, upstream pressure $p_u$ $<\cdot>$ \si{\bar}, downstream pressure $p_d$ $<\cdot>$ \si{\bar}, vapour pressure $p_v$ $<\cdot>$ \si{\bar}, volumetric flow rate $Q$ $<\cdot>$ \si{\cubic\metre\per\hour}, pressure recovery factor $X_f$ $<\cdot>$, fluid temperature $T$ $<\cdot>$ \si{\degreeCelsius}, valve flow coefficient $K_v$ $<\cdot>$ \si{\cubic\metre\per\hour} and valve stroke $h$ $<\cdot>$ \si{\milli\metre}}. For each time-frequency input, the physical operating variables are determined by its corresponding acoustic recording. The CWT time-frequency image $I_{g,n}$ and the corresponding candidate descriptions $\mathcal{T}(c,\mathbf{x}_g)$ are then encoded by the visual encoder $E_v$ and text encoder $E_t$, respectively:
\begin{equation}
\begin{aligned}
[\mathbf{g}_{g,n},\mathbf{P}_{g,n}]
    &= E_v(I_{g,n}),\\
\mathbf{t}_{g,c}
    &= E_t(\mathcal{T}(c,\mathbf{x}_g)),
\end{aligned}
\end{equation}
where $\mathbf{g}_{g,n}$ denotes the global visual representation of the current time-frequency image, $\mathbf{P}_{g,n}=[\mathbf{p}_{g,n,1},\mathbf{p}_{g,n,2},\ldots,\mathbf{p}_{g,n,N_p}]^{\mathrm T}$ represents its local patch representations and $\mathbf{t}_{g,c}$ denotes the textual feature corresponding to candidate class $c$ under operating condition $\mathbf{x}_g$. In this way, the physical operating information is incorporated into the candidate semantics before vision-language matching.

However, the text features obtained above are encoded independently of the acoustic patterns observed in the current time-frequency window. To adapt the candidate semantics according to the current acoustic response, POC-VLM introduces a time-frequency-conditioned text adaptation module, producing an adapted representation $\widetilde{\mathbf{T}}_{g,n}$ that conditions the operating-condition-aware candidate representations $\{\mathbf{t}_{g,c}\}_{c=1}^{C}$ on the time-frequency response of the current acoustic window $I_{g,n}$.
\begin{equation}
\label{eq: cross-attention}
\begin{aligned}
\mathbf{A}_{g,n}
    &= \text{Cross-Attention}(
    Q=\mathbf{T}_g,
    K=\mathbf{P}_{g,n}+\mathbf{E}_{\mathrm{pos}},
    V=\mathbf{P}_{g,n}),\\
\mathbf{U}_{g,n}
    &= \text{LayerNorm}(
    \mathbf{T}_g+\mathbf{W}_m\mathbf{A}_{g,n}),\\
\widetilde{\mathbf{T}}_{g,n}
    &= \mathbf{T}_g+\boldsymbol{\gamma}\odot\mathbf{U}_{g,n},
\end{aligned}
\end{equation}
where $\mathbf{T}_g=[\mathbf{t}_{g,1},\mathbf{t}_{g,2},\cdots,\mathbf{t}_{g,C}]^{\mathrm T}$ denotes the stacked textual representations of all candidate cavitation states, $\mathbf{E}_{\mathrm{pos}}$ denotes the positional encoding of the local time-frequency patches, $\mathbf{W}_m$ is a learnable linear projection and $\boldsymbol{\gamma}$ is a learnable residual scaling parameter. In the cross-attention operation, $\mathbf{T}_g$ provides the queries, while $\mathbf{P}_{g,n}$ serves as keys and values. This design allows each candidate state to selectively attend to the local acoustic patterns most relevant to its semantics. As a result, the adapted representation $\widetilde{\mathbf{t}}_{g,n,c}=[\widetilde{\mathbf{T}}_{g,n}]_c$ is conditioned on the observed time-frequency response, establishing a direct interaction between the operating-condition-aware candidate semantics and local acoustic evidence.

\subsection{Global-Local Vision-Language Matching}
\label{subsec: Global-Local Vision-Language Matching}
POC-VLM performs global-local vision-language matching to measure the compatibility between the current time-frequency response and each candidate cavitation state. The global visual representation $\mathbf{g}_{g,n}$, local patch representation $\mathbf{p}_{g,n,j}$ and adapted textual representation $\widetilde{\mathbf{t}}_{g,n,c}$ are first projected into a common matching space and normalized as:
\begin{equation}
\begin{aligned}
\bar{\mathbf{g}}_{g,n}
    &= \operatorname{Norm}(
    \mathbf{W}_G\mathbf{g}_{g,n}+\mathbf{b}_G),\\
\bar{\mathbf{p}}_{g,n,j}
    &= \operatorname{Norm}(
    \mathbf{W}_L\mathbf{p}_{g,n,j}+\mathbf{b}_L),\\
\bar{\mathbf{t}}_{g,n,c}
    &= \operatorname{Norm}(
    \mathbf{W}_T\widetilde{\mathbf{t}}_{g,n,c}+\mathbf{b}_T),
\end{aligned}
\end{equation}
where $\mathbf{W}_G$, $\mathbf{W}_L$ and $\mathbf{W}_T$ denote the learnable projections for the global visual, local visual and adapted textual representations, respectively. The global matching score $s_{g,n,c}^{G} = \bar{\mathbf{g}}_{g,n}^{\rm T}\bar{\mathbf{t}}_{g,n,c}$ measures the semantic consistency between the overall time-frequency response and candidate cavitation state $c$. In parallel, the patch-wise similarity $s_{g,n,c,j}^{P} = \bar{\mathbf{p}}_{g,n,j}^{\rm T}\bar{\mathbf{t}}_{g,n,c}$ measures the correspondence between the $j$-th local time-frequency patch and candidate state $c$. For each candidate state $c$, the local patches are ranked according to $s_{g,n,c,j}^{P}$ and the set of the $K_L$ highest-scoring patches is denoted by:
\begin{equation}
\Omega_{g,n,c} = \text{Top-}\text{K}_{K_L}(\{s_{g,n,c,j}^{P}\}_{j=1}^{N_p}).
\end{equation}

The corresponding local matching score is calculated as:
\begin{equation}
s_{g,n,c}^{L} = \frac{1}{K_L}\sum\limits_{j \in \Omega_{g,n,c}} s_{g,n,c,j}^{P}.
\end{equation}

Since $\Omega_{g,n,c}$ is determined independently for each candidate class, different cavitation states can be associated with different discriminative regions of the same time-frequency representation. Thus, $s_{g,n,c}^{G}$ characterizes the overall acoustic response, whereas $s_{g,n,c}^{L}$ provides complementary class-specific evidence from localized time-frequency regions. Finally, the global and local matching scores are combined as:
\begin{equation}
z_{g,n,c}^{\text{VL}} = \tau (\alpha_G s_{g,n,c}^{G} + \alpha_L s_{g,n,c}^{L}),
\end{equation}
where $\alpha_G$ and $\alpha_L$ denote the normalized learnable weights of the global and local matching terms, respectively, and $\tau$ is a learnable scaling parameter. The resulting vision-language score $z_{g,n,c}^{\text{VL}}$ integrates the overall time-frequency response and candidate-specific local acoustic evidence for subsequent fusion with the physical operating-condition prior.

\subsection{Physical Operating-Condition Prior and Evidence Fusion}
\label{subsec: Physical Operating-Condition Prior and Evidence Fusion}
POC-VLM constructs a physical operating-condition prior from labeled training recordings, representing each recording by one prototype. The selected physical variables are standardized using training-prototype statistics before nearest-neighbour estimation. For the current operating condition $\hat{\mathbf{x}}_g$, the distance $d_{g,i} = \left\| \hat{\mathbf{x}}_g - \hat{\mathbf{x}}_i \right\|_2$ to the $i$-th training prototype $\hat{\mathbf{x}}_i$ is calculated in the standardized physical operating-condition space. Accordingly, the physical operating-condition prior $P_{\text{POC}}(c|\mathbf{x}_g)$ for candidate class $c$ is estimated by inverse-distance weighted voting:
\begin{equation}
P_{\text{POC}}(c|\mathbf{x}_g) = \frac{{\sum\nolimits_{i \in {\mathcal{N}_k}(\mathbf{x}_g)} {w_{g,i}\mathbb{I}(y_i = c)} }}{{\sum\nolimits_{i \in {\mathcal{N}_k}(\mathbf{x}_g)} {w_{g,i}} }},
\end{equation}
where ${\mathcal{N}_k}(\mathbf{x}_g)$ denotes the set of the $k$ nearest training-condition prototypes, $\mathbb{I}(\cdot)$ denotes the indicator function, $w_{g,i} = 1/\max (d_{g,i},\varepsilon )$ denotes the inverse-distance weight assigned to the $i$-th neighbouring prototype and $\varepsilon$ is a small constant introduced for numerical stability. The resulting physical operating-condition prior $P_{\text{POC}}(c|\mathbf{x}_g)$ is then fused with the vision-language matching score $z_{g,n,c}^{\text{VL}}$ at the logit level to obtain the final diagnostic score $z_{g,n,c}^{\text{final}}$, as defined in Eq. \ref{eq: final score}.

\section{Experiment Setup}
\label{sec: 4_ExperimentSetup}
\subsection{Industrial Cavitation Experimental Platform}
\label{subsec: Industrial Cavitation Experimental Platform}
\begin{figure}[htbp]
\centering
\includegraphics[width=0.9\textwidth,height=40mm]{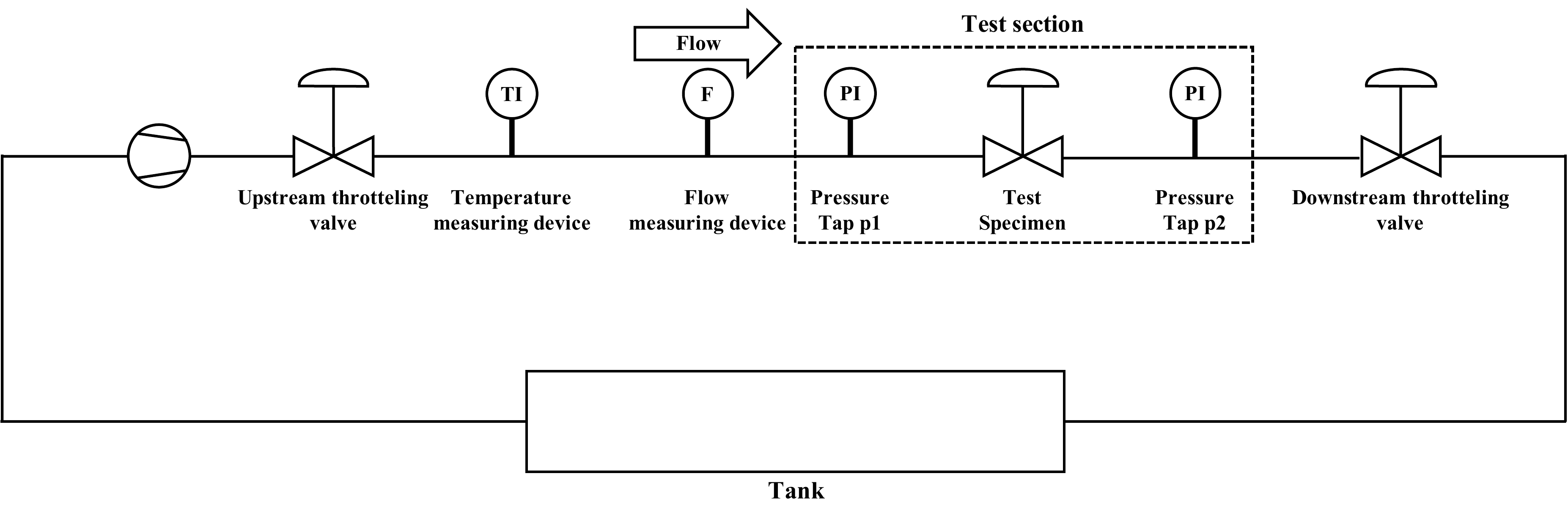}
\caption{Schematic view of the test rack at SAMSON AG.}
\label{fig: system}
\end{figure}

The cavitation experiments were conducted on an industrial control-valve test platform developed by SAMSON AG, as shown in Figure \ref{fig: system}. The test section contains a SAMSON Type 3241 control valve (DN 80, PN 40, $K_{vs}=$25) equipped with a Type 3730-6 positioner, with heated water used as the working medium. The platform measures the upstream pressure $p_u$ (\si{\bar\,(g)}), downstream pressure $p_d$ (\si{\bar\,(g)}), vapor pressure $p_v$ (\si{\bar\,(g)}), volumetric flow rate $Q$ (\si{\cubic\metre\per\hour}), fluid temperature $T$ (\si{\degreeCelsius}), valve flow coefficient $K_v$ (\si{\cubic\metre\per\hour}) and valve stroke $h$ (\si{\milli\metre}). Based on the measured pressures, the pressure recovery factor ${X_f}$ is calculated as ${{({p_u} - {p_d})} \mathord{\left/{\vphantom {{({p_u} - {p_d})} {({p_u} - {p_v})}}} \right.\kern-\nulldelimiterspace} {({p_u} - {p_v})}}$. Two external throttling control valves are installed upstream and downstream of the test valve, while the pump and throttling valves jointly regulate the flow and pressure conditions. The pipe sections between the external throttling valves and the test valve are sufficiently long to provide an undisturbed flow field for reliable measurements. The detailed specifications of the sensors used in the platform are summarized in Table \ref{tab: sensors}.
\begin{table}[htbp]
\caption{Detailed information of various measurement sensors.}
\label{tab: sensors}
\centering
\footnotesize
\setlength{\tabcolsep}{7.6mm}{
\begin{tabular}{llll}
\toprule
Physical parameters          & Types             & Ranges                                & Tolerances \\ 
\midrule
Flow rate                    & Krohne  M950      & 0$-$\qty{180}{\cubic\m\per\hour}      & $\pm$ \qty{0.9}{\cubic\m\per\hour}\\
Valve stroke                 & Sylvac  S229      & 0$-$\qty{50}{\mm}                     & $\pm$ \qty{0.001}{\mm}\\
Upstream pressure            & SAMSON  6054      & 0$-$\qty{16}{\bar}                    & $\pm$ \qty{0.09}{\bar} \\
Downstream pressure          & SAMSON  6054      & 0$-$\qty{16}{\bar}                    & $\pm$ \qty{0.09}{\bar} \\
Medium temperature           & SAMSON  5204      & -20$-$\qty{150}{\degreeCelsius}       & $\pm$ \qty{0.45}{\degreeCelsius}  \\
Valve structure noise        & Vallen  VS45-H    & 20$-$\qty{450}{\kHz}                  & $\pm$ \qty{63}{\dB}\\
Valve structure noise        & PCB  M353 B17     & 0$-$\qty{30}{\kHz}                    & $\pm$ \qty{3}{\dB}    \\
Bonnet structure noise       & PCB  M353 B17     & 0$-$\qty{30}{\kHz}                    & $\pm$ \qty{3}{\dB}    \\ 
\bottomrule
\end{tabular}}
\end{table}

To capture cavitation-related acoustic responses from different sensing perspectives, two piezoelectric sensors and two microphones are employed. One piezoelectric sensor is mounted directly on the control-valve body, while the other is installed on the NAMUR\footnote{Nationaler Arbeitskreis f\"ur Mess- und Regeltechnik in der Chemischen Industrie} interface attached to the valve bonnet to capture structure-borne acoustic responses. In addition, a high-end microphone and a low-cost microphone typically used in smartphones are positioned approximately \SI{1}{\m} from the control valve to record airborne acoustic signals over the frequency range \SI{40}{\Hz} and \SI{20}{\kHz}. During all experiments, the water temperature is maintained between \SI{25}{\degreeCelsius} $-$ \SI{40}{\degreeCelsius} to limit variations in vapour pressure.

The test section is located in a separate room from major external noise sources, including the pump and the upstream and downstream throttling valves, which are installed beneath the test section. Expansion joints and bellows expansion joints are incorporated into the pipework beneath the test rack to manage vibration transmission through the piping system. Together with the spatial separation of the test section, these arrangements keep the ambient noise level below approximately \SI{55}{\dB}. All experiments are conducted in accordance with IEC 60534-8-2. By adjusting the valve opening together with the upstream and downstream pressure conditions, different cavitation states are generated under controlled physical operating conditions for subsequent acoustic diagnosis.

\subsection{Datasets and Data Preprocessing}
\label{subsec: Datasets and Data Preprocessing}
Three real-world industrial cavitation datasets collected by SAMSON AG, namely \textbf{Cavitation-Short}, \textbf{Cavitation-Long} and \textbf{Cavitation-Noise} are used for evaluation. Cavitation-Short consists of \SI{3}{\s} acoustic recordings, whereas Cavitation-Long and Cavitation-Noise contain \SI{25}{\s} recordings. Cavitation-Noise further incorporates environmental noise to represent more challenging operating conditions. All acoustic signals are sampled at \SI{1562.5}{\kHz}. Following the diagnostic formulation in Section \ref{subsec: Cavitation Intensity Diagnosis}, turbulent-flow and no-flow conditions are grouped into the non-cavitation class. The operating conditions and the detailed training/test distributions of the three datasets are summarized in Tables \ref{tab: CavitationDatasets-operation} and \ref{tab: training and test sets}, respectively.
\begin{table}[htbp]
\centering
\footnotesize
\caption{Details of three real-world cavitation datasets for valve operation with various upstream pressures.}
\label{tab: CavitationDatasets-operation}

\setlength{\tabcolsep}{3.5mm}{
\begin{tabular}{lccc}
\toprule 
\multicolumn{1}{c}{\multirow{2}{*}{Dataset}} 
& \multicolumn{3}{c}{Operation parameters} \\ 
\cmidrule{2-4} 
& Valve stroke (\si{\mm})
& Upstream pressure (\si{\bar})
& Temperature (\si{\degreeCelsius}) \\ 
\midrule

Cavitation-Short 
& {[}15, 13.5, 11.25, 7.5, 3.75, 1.5, 0.75{]}  
& {[}10, 9, 6, 4{]} 
& 25--50 \\

Cavitation-Long 
& {[}60, 55, 45, 30, 25, 15, 6{]} 
& {[}10, 6, 4{]} 
& 23--52 \\

Cavitation-Noise 
& 15 
& 10 
& 32--39 \\ 

\bottomrule
\end{tabular}
}
\end{table}
\begin{table}[htbp]
\caption{Details of the training and test sets. $(\cdot )$ denotes the number after the sliding window (a window duration of \SI{0.06}{s},  an overlap ratio of 0.25 and 512 CWT scales).}
\label{tab: training and test sets}
\footnotesize
\centering
\setlength{\tabcolsep}{1.2mm}{
\begin{tabular}{lcccccccc}
\toprule
                 \multicolumn{1}{c}{\multirow{3}{*}{Dataset}} & \multicolumn{4}{c}{Training set}      & \multicolumn{4}{c}{Testing set} \\ 
\cmidrule{2-9} 
                 & \multicolumn{3}{c}{Cavitation}       
                 & \multicolumn{1}{c|}{\multirow{2}{*}{Non}}
                 & \multicolumn{3}{c}{Cavitation}     
                 & \multirow{2}{*}{Non} \\ 
\cmidrule{2-4}
\cmidrule{6-8}
                 & Choked flow & Constant   & Incipient & \multicolumn{1}{c|}{}     & Choked flow & Constant & Incipient &    \\ 
\midrule
Cavitation-Short & 58($\times$66)    & 75($\times$66)   & 32($\times$66)  & \multicolumn{1}{c|}{121($\times$66)}  & 14($\times$66)    & 18($\times$66) & 8($\times$66)    & 30($\times$66)    \\
Cavitation-Long  & 117($\times$555)   & 315($\times$555) & 55($\times$555)  & \multicolumn{1}{c|}{158($\times$555)}  & 31($\times$555)    & 81($\times$555) & 9($\times$555)  & 40($\times$555)    \\
Cavitation-Noise & 32($\times$555)    & 32($\times$555)   & 32($\times$555)  & \multicolumn{1}{c|}{32($\times$555)}    & 8($\times$555)      & 8($\times$555)   & 8($\times$555)    & 8($\times$555)      \\ 
\bottomrule
\end{tabular}
}
\end{table}

The original acoustic recordings are divided into training and test sets before window segmentation to prevent windows from the same recording from appearing in different subsets. Each recording is subsequently divided into fixed-duration subsequences using a sliding window with a duration of \SI{0.06}{\s} and an overlap ratio of 0.25. Given an acoustic signal, the window size and step size are defined as $W_s=\text{round}(0.06f_s)$ and $S=\text{round}[W_s(1-0.25)]$, respectively. The resulting numbers of training and test samples after segmentation are also reported in Table \ref{tab: training and test sets}. For each acoustic subsequence, the CWT is applied to obtain its time-frequency representation. A real-valued Morlet wavelet with 512 scales is employed to capture the transient and non-stationary characteristics of cavitation acoustic responses. The resulting time-frequency representations are converted into RGB spectrograms and used as the visual inputs of POC-VLM.

\subsection{Comparison Methods}
\label{subsec: Comparison Methods}
To comprehensively evaluate POC-VLM, the comparison methods are organized into four groups according to the information used and their modelling strategies.

\textbf{Data-driven methods.} This group directly learns diagnostic representations from acoustic time-frequency inputs without explicitly incorporating physical operating-condition variables. It includes CNN-based methods such as ResNet \cite{he2016deep}, DenseNet \cite{huang2017densely} and ConvNeXt \cite{liu2022convnet}. Transformer-based methods include ViT \cite{dosovitskiy2020image}, Swin \cite{liu2021swin}, PerViT \cite{min2022peripheral} and CrossViT \cite{chen2021crossvit}, while CNN-Transformer hybrid architectures include UniFormer \cite{li2023uniformer}, ConFormer \cite{gulati2020conformer}, CFFormer \cite{li2026cfformer} and LeViT \cite{graham2021levit}. These methods provide representative baselines for evaluating the diagnostic capability of purely data-driven representation learning.

\textbf{Physical-variable-driven methods.} Traditional machine-learning methods using physical operating-condition variables as input features are also included for comparison, including support vector machine (SVM) \cite{cortes1995support}, decision tree (DT) \cite{quinlan1986induction}, extreme gradient boosting (XGBoost) \cite{chen2016xgboost} and K-nearest neighbours (KNN) \cite{fukunaga2013introduction}. These baselines evaluate the discriminative capability of physical operating conditions independently of acoustic time-frequency representations.

\textbf{Data-physics hybrid methods.} This group includes representative multimodal learning and vision-language adaptation approaches for exploiting complementary information from acoustic observations and physical operating conditions. Specifically, ViLT \cite{kim2021vilt}, Perceiver \cite{jaegle2021perceiver}, PerceiverIO \cite{jaegle2021perceiverio} and MulT \cite{tsai2019multimodal} are adopted as multimodal baselines, using different fusion and attention mechanisms to model interactions between heterogeneous inputs. In addition, CoOp \cite{zhou2022learning}, CoCoOp \cite{zhou2022conditional}, CLIP-Adapter \cite{gao2024clip}, Tip-Adapter \cite{zhang2022tip}, TaskRes \cite{yu2023task} and MaPLe \cite{khattak2023maple} are included as representative vision-language adaptation methods, covering prompt learning, feature adaptation and residual-based adaptation strategies. Together, these methods provide a broader comparison with POC-VLM in terms of multimodal information fusion, cross-modal semantic alignment and task-specific adaptation.

\textbf{Domain-specific fault intensity diagnosis methods.} In addition to the above general architectures, several methods specifically developed for fault or cavitation intensity diagnosis are considered, including LiftingNet \cite{pan2017liftingnet}, MIPLCNet \cite{pan2019novel}, ResNet-APReLU \cite{zhao2020deep}, LSTM-RDRN \cite{mohammad2023one}, HKG-ResNet34 \cite{sha2024hierarchical}, HKG-Swin-B \cite{sha2024hierarchical}, DHK+ConvNeXt-B \cite{sha2026deep} and DHK+Swin-B \cite{sha2026deep}. These methods provide more task-specific comparisons for evaluating POC-VLM against existing fault-intensity diagnostic approaches.

\subsection{Implementation Details}
\label{subsec: Implementation Details}
All experiments are implemented in Python 3.10.19 using PyTorch 2.8.0 with CUDA 12.9 and cuDNN 9.10.2 on a Linux workstation equipped with two NVIDIA GeForce RTX 5090 GPUs. Each training run uses a single GPU, and the same model configuration is adopted for all three datasets. The input RGB time-frequency spectrograms are resized to $224 \times 224$. POC-VLM employs the pretrained CLIP ViT-B/16 \cite{radford2021learning,liu2023visual} as the vision-language backbone, with both the visual encoder $E_v$ and text encoder $E_t$ kept frozen throughout training. The visual encoder produces a 512-dimensional global representation and 196 local patch representations of the same dimension. The P2Tformer contains one cross-attention layer with eight attention heads and a feature dimension of 512 \cite{li2025fgaseg}. The residual scaling parameter $\gamma$ is initialized to 0.1. For global-local vision-language matching, the local branch selects the ${K_L} = 16$ highest-scoring patches for each candidate class. The normalized global and local matching weights $\alpha_G$ and $\alpha_L$ are both initialized to 0.5, while the matching scale $\tau$ is initialized to $1/0.07$. Candidate textual descriptions are tokenized using the CLIP tokenizer with a context length of 77 tokens.

Only the P2Tformer and global-local matching modules are optimized, while $E_v$ and $E_t$ remain fixed and the physical operating-condition prior contains no learnable parameters. The trainable parameters are optimized using AdamW, and the model is trained for 30 epochs with a batch size of 32. Weighted label-smoothed cross-entropy is adopted as the training objective. Given the predicted probability $p_{b,c}$ of sample $b$ belonging to class $c$, the classification loss is defined as:
\begin{equation}
\mathcal{L} = \frac{{\sum\nolimits_{b = 1}^B {[(1 - \varepsilon ){w_{yb}}( - \log {p_{b,yb}}) + \frac{\varepsilon }{C}\sum\nolimits_{c = 1}^C {{w_c}( - \log {p_{b,c}})} ]} }}{{\sum\nolimits_{b = 1}^B {{w_{yb}}} }},
\end{equation}
where $C=4$ denotes the number of cavitation classes, $\varepsilon=0.02$ is the label-smoothing factor and $w_c$ denotes the weight assigned to class $c$. The class weights $w_c$ are calculated separately from the class frequencies of each training set using inverse square-root frequency weighting and are normalized to have a mean of one. Full-model gradient clipping with a maximum $L_2$ norm of 1.0 is applied during optimization.

For the physical operating-condition prior, the KNN prototype library is constructed exclusively from the training recordings, with one prototype retained for each original operating condition. The same physical operating-condition variables constituting and the same KNN configuration are used for Cavitation-Short, Cavitation-Long and Cavitation-Noise. For each dataset, the selected variables are standardized using the mean and population standard deviation calculated from the corresponding training prototypes. The $k=3$ nearest prototypes are selected and inverse-distance weighted voting is used to estimate the physical operating-condition prior. The numerical stability constant is set to $\delta  = {10^{ - 6}}$ and the fusion coefficient $\lambda_{\text{POC}}=1$ for all three datasets.

\subsection{Evaluation Metrics}
\label{subsec: Evaluation Metrics}
The diagnostic performance is evaluated using Accuracy (Acc.), Precision (Pre.), Recall (Rec.), F1-score (F1) and the area under the receiver operating characteristic curve (AUC-ROC). Accuracy measures the overall proportion of correctly classified samples. Precision, Recall and F1-score are reported as macro averages, giving equal weight to each cavitation class regardless of its frequency. The metric definitions are provided in Appendix \ref{app: Evaluation Metrics}. Higher values of all metrics indicate better diagnostic performance.

\section{Experimental Result and Analysis}
\label{sec: 5_ExperimentalResultAnalysis}

\subsection{Diagnostic Performance}
\label{subsec: Diagnostic Performance}
Table \ref{tab: OverallResults} compares POC-VLM with representative data-driven, physical-variable-driven, multimodal and CLIP-based diagnostic methods across the three real-world cavitation datasets. Overall, POC-VLM achieves the best performance on all three datasets. On Cavitation-Short, POC-VLM obtains an accuracy of 97.14$\%$, together with a precision, recall and F1-score of 96.94$\%$, 95.49$\%$ and 96.08$\%$, respectively. Compared with MulT, the strongest competing method on this dataset, POC-VLM improves the accuracy and F1-score by 5.15$\%$ and 4.76$\%$, respectively. On Cavitation-Long, POC-VLM achieves an accuracy of 98.76$\%$ and an F1-score of 97.81$\%$, exceeding PerceiverIO by 4.73$\%$ in accuracy and Perceiver by 3.89$\%$ in F1-score. On Cavitation-Noise, POC-VLM reaches 100$\%$ on all four evaluation metrics, improving the best competing accuracy and F1-score of 99.40$\%$ by 0.60$\%$.

Notably, POC-VLM also consistently outperforms the CLIP-based adaptation methods, including CoOp, CoCoOp, CLIP-Adapter, Tip-Adapter, TaskRes and MaPLe. Among these methods, MaPLe achieves the highest accuracies of 88.57$\%$, 93.17$\%$ and 98.99$\%$ on Cavitation-Short, Cavitation-Long and Cavitation-Noise, respectively, whereas POC-VLM further improves them by 8.57$\%$, 5.59$\%$ and 1.01$\%$. These results suggest that directly adapting pretrained vision-language representations alone is insufficient for condition-dependent cavitation diagnosis. Explicitly incorporating sample-specific physical operating conditions into both semantic representation and decision-level reasoning provides additional diagnostic information beyond conventional data-driven, multimodal and CLIP-based adaptation strategies.
\begin{table}[htbp]

\caption{Diagnostic performance comparison of different methods across real-world industrial cavitation datasets.}

\label{tab: OverallResults}

\footnotesize

\centering

\setlength{\tabcolsep}{1.3mm}

\begin{tabular}{c|c|cccc|cccc|cccc}

\toprule

\multirow{2}{*}{Types} &
\multirow{2}{*}{Methods}
& \multicolumn{4}{c|}{Cavitation-Short}
& \multicolumn{4}{c|}{Cavitation-Long}
& \multicolumn{4}{c}{Cavitation-Noise}
\\

\cmidrule(lr){3-6}
\cmidrule(lr){7-10}
\cmidrule(lr){11-14}

&
& Acc. & Pre. & Rec. & F1
& Acc. & Pre. & Rec. & F1
& Acc. & Pre. & Rec. & F1
\\

\midrule

\multirow{29}{*}{Data}

& ResNet18
& 89.39 & 89.40 & 89.39 & 88.51
& 93.35 & 93.33 & 93.35 & 93.33
& 98.05 & 98.05 & 98.05 & 98.05 \\

& ResNet34
& 89.55 & 89.71 & 89.55 & 88.79
& 93.38 & 93.30 & 93.38 & 93.32
& 98.06 & 98.06 & 98.06 & 98.06 \\

& ResNet50
& 89.24 & 89.24 & 89.24 & 88.48
& \underline{93.44} & \underline{93.35} & \underline{93.44} & \underline{93.36}
& 97.84 & 97.84 & 97.84 & 97.84 \\

& DenseNet121
& \underline{90.11} & \underline{90.22} & \underline{90.11} & \underline{89.45}
& 93.30 & 93.17 & 93.30 & 93.19
& \underline{98.58} & \underline{98.58} & \underline{98.58} & \underline{98.58} \\

& DenseNet161
& 89.94 & 89.99 & 89.94 & 89.23
& 93.40 & 93.23 & 93.40 & 93.27
& 97.76 & 97.76 & 97.76 & 97.76 \\

& ConvNeXt-T
& 87.60 & 87.20 & 87.60 & 87.04
& 92.93 & 92.86 & 92.93 & 92.87
& 97.96 & 97.96 & 97.96 & 97.96 \\

& ConvNeXt-S
& 87.45 & 87.71 & 87.45 & 86.11
& 92.67 & 92.70 & 92.67 & 92.66
& 98.48 & 98.48 & 98.48 & 98.48 \\

& ConvNeXt-B
& 88.20 & 88.07 & 88.20 & 87.37
& 92.62 & 92.61 & 92.62 & 92.59
& 97.80 & 97.80 & 97.80 & 97.80 \\

& ConvNeXt-L
& 87.99 & 88.14 & 87.99 & 87.37
& 93.12 & 93.00 & 93.12 & 93.04
& 97.48 & 97.48 & 97.48 & 97.48 \\

\cmidrule{2-14}

& ViT-T
& 89.59 & 90.04 & 89.59 & 88.66
& 92.21 & 92.04 & 92.21 & 92.11
& 98.33 & 98.33 & 98.33 & 98.33 \\

& ViT-S
& 88.90 & 88.92 & 88.90 & 87.70
& 92.53 & 92.50 & 92.53 & 92.49
& 98.27 & 98.27 & 98.27 & 98.27 \\

& ViT-B
& 89.26 & 89.80 & 89.26 & 88.24
& 92.23 & 91.99 & 92.23 & 92.09
& 97.80 & 97.80 & 97.80 & 97.80 \\

& Swin-T
& 89.37 & 89.47 & 89.37 & 89.07
& 93.41 & 93.24 & 93.41 & 93.26
& 98.64 & 98.64 & 98.64 & 98.64 \\

& Swin-S
& 89.31 & 89.04 & 89.31 & 88.90
& 92.61 & 92.45 & 92.61 & 92.46
& 98.40 & 98.40 & 98.40 & 98.40 \\

& Swin-B
& \underline{90.37} & 90.49 & \underline{90.37} & \underline{89.71}
& 91.46 & 91.10 & 91.46 & 91.20
& 97.76 & 97.76 & 97.76 & 97.76 \\

& PerViT-T
& 90.26 & \underline{90.54} & 90.26 & 89.53
& \underline{93.90} & \underline{93.78} & \underline{93.90} & \underline{93.82}
& \underline{98.68} & \underline{98.68} & \underline{98.68} & \underline{98.68} \\

& PerViT-S
& 88.70 & 89.58 & 88.70 & 86.76
& 92.76 & 92.76 & 92.76 & 92.71
& 98.50 & 98.50 & 98.50 & 98.50 \\

& PerViT-M
& 88.78 & 88.71 & 88.68 & 87.79
& 93.20 & 92.97 & 93.20 & 93.04
& 97.86 & 97.86 & 97.86 & 97.86 \\

& CrossViT-T
& 89.46 & 89.91 & 89.46 & 88.39
& 92.25 & 91.96 & 92.25 & 92.06
& 98.51 & 98.51 & 98.51 & 98.51 \\

& CrossViT-S
& 89.87 & 89.75 & 89.87 & 89.04
& 92.48 & 92.35 & 92.48 & 92.41
& 98.63 & 98.63 & 98.63 & 98.63 \\

& CrossViT-B
& 89.76 & 89.86 & 89.76 & 88.89
& 92.31 & 92.31 & 92.31 & 92.26
& 98.31 & 98.31 & 98.31 & 98.31 \\

\cmidrule{2-14}

& UniFormer-S
& 88.81 & 89.45 & 88.81 & 88.01
& 93.22 & 93.13 & 93.22 & 93.16
& \underline{98.71} & \underline{98.71} & \underline{98.71} & \underline{98.71} \\

& UniFormer-B
& 88.44 & 89.17 & 88.44 & 87.58
& 93.51 & 93.33 & 93.51 & 93.39
& 98.64 & 98.64 & 98.64 & 98.64 \\

& ConFormer
& 89.33 & \underline{90.14} & 89.33 & 88.33
& 93.03 & 92.85 & 93.03 & 92.91
& 98.68 & 98.68 & 98.68 & 98.68 \\

& CFFormer-S
& 89.24 & 89.47 & 89.24 & 88.53
& 93.53 & 93.48 & 93.53 & 93.45
& 98.62 & 98.62 & 98.62 & 98.62 \\

& CFFormer-B
& 88.87 & 88.86 & 88.87 & 88.29
& \underline{93.60} & \underline{93.53} & \underline{93.60} & \underline{93.54}
& 98.61 & 98.61 & 98.61 & 98.61 \\

& CFFormer-L
& \underline{89.68} & 89.67 & \underline{89.68} & \underline{88.97}
& 93.55 & 93.46 & 93.55 & 93.48
& 97.65 & 97.65 & 97.65 & 97.65 \\

& LeViT-S
& 88.83 & 88.61 & 88.83 & 87.98
& 93.05 & 92.95 & 93.05 & 92.97
& 98.63 & 98.63 & 98.63 & 98.63 \\

& LeViT-B
& 89.11 & 88.93 & 89.11 & 88.56
& 93.09 & 93.00 & 93.09 & 93.03
& 98.54 & 98.54 & 98.54 & 98.54 \\

\midrule

\multirow{4}{*}{Physics}

& SVM
& 88.57 & 82.77 & 80.76 & 81.07
& 90.96 & \underline{90.71} & \underline{90.96} & \underline{90.69}
& 99.18 & 99.18 & 99.18 & 99.18 \\

& DT
& 85.71 & 81.21 & 81.62 & 81.37
& 88.20 & 81.15 & 78.01 & 78.51
& 98.33 & 98.33 & 98.33 & 98.33 \\

& KNN
& 88.57 & 91.24 & \underline{82.10} & \underline{84.52}
& 91.93 & 90.56 & 89.60 & 89.80
& 98.47 & 98.47 & 98.47 & 98.47 \\

& XGBoost
& \underline{90.00} & \underline{92.11} & 81.59 & 83.21
& \underline{93.17} & 89.09 & 87.70 & 88.30
& \underline{99.36} & \underline{99.36} & \underline{99.36} & \underline{99.36} \\

\midrule

\multirow{11}{*}{Hybrid}

& ViLT
& 90.63 & 91.58 & 90.63 & 89.39
& 92.10 & 92.09 & 92.10 & 92.01
& 99.19 & 99.20 & 99.19 & 99.19 \\

& Perceiver
& 91.21 & 90.99 & 91.21 & 90.69
& 93.98 & \underline{93.90} & 93.98 & \underline{93.92}
& 99.27 & 99.27 & 99.27 & 99.27 \\

& PerceiverIO
& 91.49 & 91.26 & 91.49 & 90.94
& \underline{94.03} & 93.80 & \underline{94.03} & 93.85
& \underline{99.40} & \underline{99.40} & \underline{99.40} & \underline{99.40} \\

& MulT
& \underline{91.99} & \underline{92.51} & \underline{91.99} & \underline{91.32}
& 93.55 & 93.40 & 93.55 & 93.43
& 98.33 & 98.33 & 98.33 & 98.33 \\

& CoOp
& 81.43 & 89.55 & 68.20 & 69.54
& 82.61 & 73.33 & 78.54 & 74.95
& 92.90 & 92.72 & 92.90 & 92.78 \\

& CoCoOp
& 81.43 & 77.51 & 68.20 & 69.29
& 85.09 & 76.33 & 83.74 & 78.14
& 93.38 & 93.29 & 93.38 & 93.32 \\

& CLIP-Adapter
& 80.00 & 87.49 & 67.37 & 68.36
& 84.47 & 74.84 & 80.96 & 76.60
& 92.64 & 92.54 & 92.64 & 92.57 \\

& Tip-Adapter
& 82.86 & 75.70 & 65.95 & 69.56
& 80.31 & 72.58 & 75.00 & 76.74
& 93.60 & 93.48 & 93.60 & 93.49 \\

& TaskRes
& 87.14 & 74.10 & 76.07 & 74.03
& 85.78 & 78.69 & 81.08 & 78.93
& 96.24 & 96.73 & 96.24 & 96.22 \\

& MaPLe
& 88.57 & 88.79 & 81.71 & 84.20
& 93.17 & 86.73 & 83.94 & 84.71
& 98.99 & 99.03 & 98.99 & 98.99 \\

\rowcolor{gray!15}
\cellcolor{white}

& \textbf{POC-VLM}
& \textbf{97.14} & \textbf{96.94} & \textbf{95.49} & \textbf{96.08}
& \textbf{98.76} & \textbf{99.40} & \textbf{96.42} & \textbf{97.81}
& \textbf{100.0} & \textbf{100.0} & \textbf{100.0} & \textbf{100.0} \\

\bottomrule

\end{tabular}

\end{table}

Table \ref{tab:Comparsion_cavitation_domain_specific} further compares POC-VLM with domain-specific fault and cavitation intensity diagnosis methods. POC-VLM consistently outperforms all compared task-specific approaches across the three datasets. On Cavitation-Short, POC-VLM improves the best competing accuracy from 92.14$\%$ to 97.14$\%$, corresponding to a gain of 5.00$\%$. It also improves the best precision, recall and F1-score by 4.59$\%$, 3.26$\%$ and 4.85$\%$, respectively. On Cavitation-Long, POC-VLM achieves an accuracy of 98.76$\%$, improving the strongest baseline by 4.41$\%$, while the precision, recall and F1-score are improved by 8.37$\%$, 1.99$\%$ and 5.29$\%$, respectively. On Cavitation-Noise, where the domain-specific methods already achieve near-saturated performance, POC-VLM further increases all four metrics from the best baseline value of 99.81$\%$ to 100$\%$, corresponding to an improvement of 0.19$\%$. These results indicate that the proposed framework provides consistent advantages over existing domain-specific approaches, particularly on Cavitation-Short and Cavitation-Long. Unlike methods that mainly rely on hierarchical feature learning or task-specific representation enhancement, POC-VLM jointly exploits semantic consistency between acoustic responses and physical operating-condition-aware state descriptions together with an explicit physical operating-condition prior.
\begin{table}[htbp]

\caption{Diagnostic performance comparison with domain-specific methods across real-world cavitation datasets.}

\label{tab:Comparsion_cavitation_domain_specific}

\centering

\footnotesize

\setlength{\tabcolsep}{1.1mm}

\begin{tabular}{c|c|cccc|cccc|cccc}

\toprule

\multirow{2}{*}{Types} &
\multirow{2}{*}{Methods}
& \multicolumn{4}{c|}{Cavitation-Short}
& \multicolumn{4}{c|}{Cavitation-Long}
& \multicolumn{4}{c}{Cavitation-Noise}
\\

\cmidrule(lr){3-6}
\cmidrule(lr){7-10}
\cmidrule(lr){11-14}

&
& Acc. & Pre. & Rec. & F1
& Acc. & Pre. & Rec. & F1
& Acc. & Pre. & Rec. & F1
\\

\midrule

\multirow{4}{*}{Data}

& LiftingNet
& 85.29 & 82.75 & 75.89 & 73.95
& 88.43 & 86.31 & 87.97 & 87.05
& 95.86 & 94.62 & 95.84 & 95.20 \\

& MIPLCNet
& 86.57 & 83.38 & 77.71 & 75.00
& 89.14 & 87.06 & 88.18 & 87.55
& 96.57 & 95.78 & 96.61 & 96.18 \\

& ResNet-APReLU
& 86.86 & 84.04 & 77.89 & 75.56
& 90.71 & 88.83 & 90.91 & 89.70
& 97.29 & 96.77 & 97.49 & 97.12 \\

& LSTM-RDRN
& \underline{87.71} & \underline{85.42} & \underline{78.50} & \underline{76.72}
& \underline{91.14} & \underline{89.36} & \underline{91.55} & \underline{90.32}
& \underline{98.71} & \underline{98.40} & \underline{98.53} & \underline{98.46} \\

\midrule

\multirow{5}{*}{Hybrid}

& HKG-ResNet34
& 89.71 & 90.54 & 78.02 & 76.74
& 92.44 & 88.55 & 92.21 & 90.16
& 99.06 & 99.06 & 99.06 & 99.06 \\

& HKG-Swin-B
& 89.57 & \underline{92.35} & 77.88 & 76.66
& 93.18 & 89.58 & 93.40 & 91.27
& 99.63 & 99.63 & 99.63 & 99.62 \\

& DHK+CoNeXt-B
& \underline{92.14} & 90.71 & \underline{91.80} & \underline{91.23}
& \underline{94.35} & \underline{91.03} & 94.30 & \underline{92.52}
& \underline{99.81} & \underline{99.81} & \underline{99.81} & \underline{99.81} \\

& DHK+Swin-B
& 91.86 & 89.87 & 92.23 & 90.94
& 94.19 & 90.88 & \underline{94.43} & 92.45
& 99.75 & 99.75 & 99.75 & 99.75 \\

\rowcolor{gray!15}
\cellcolor{white}

& \textbf{POC-VLM}
& \textbf{97.14} & \textbf{96.94} & \textbf{95.49} & \textbf{96.08}
& \textbf{98.76} & \textbf{99.40} & \textbf{96.42} & \textbf{97.81}
& \textbf{100.0} & \textbf{100.0} & \textbf{100.0} & \textbf{100.0} \\

\bottomrule

\end{tabular}

\end{table}

\begin{figure*}[htbp]
    \centering
    \includegraphics[width=\textwidth,height=90mm]{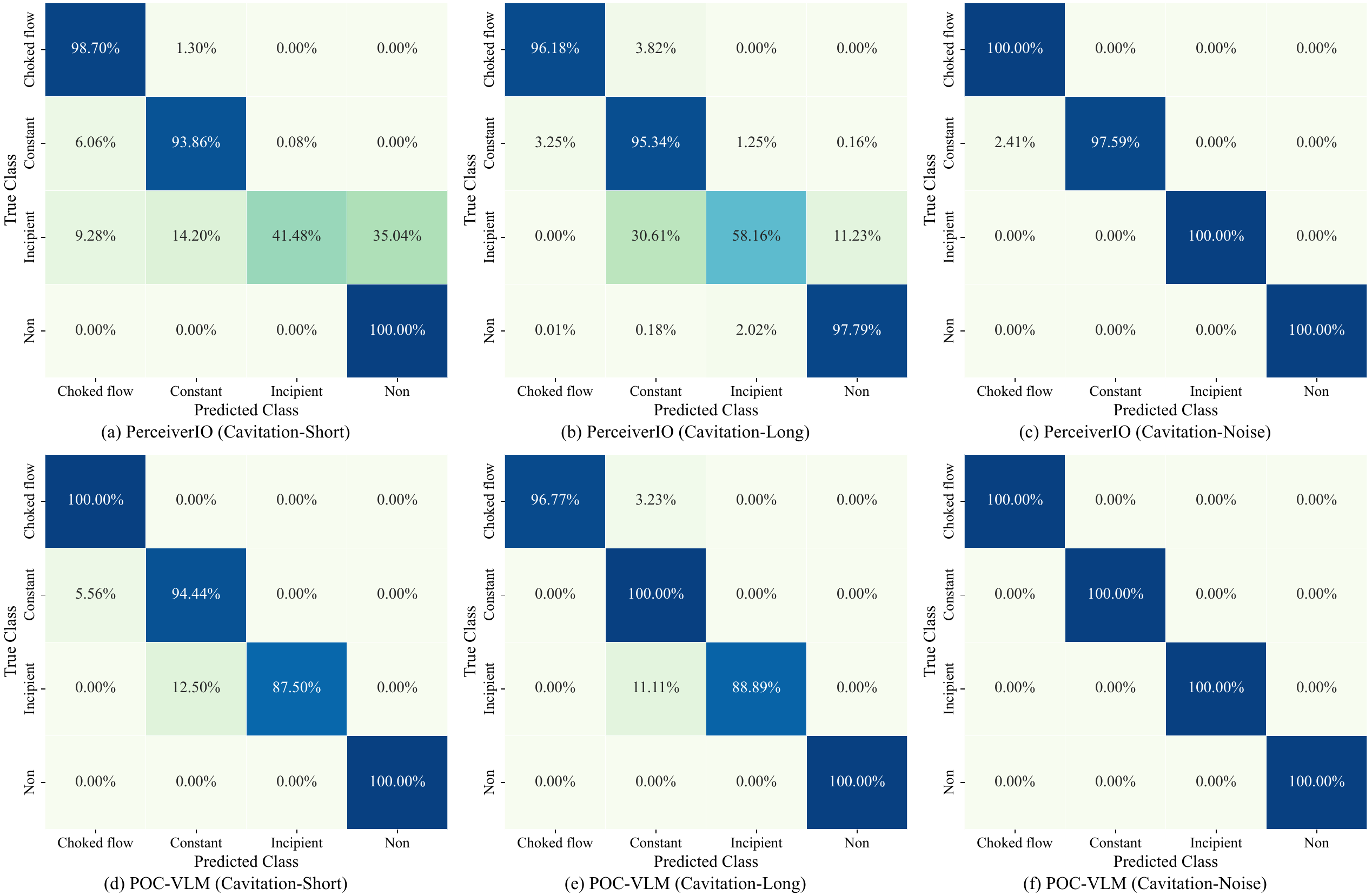}
    \caption{Normalized confusion matrices of PerceiverIO and POC-VLM across the three real-world cavitation datasets. (a-c) PerceiverIO on Cavitation-Short, Cavitation-Long and Cavitation-Noise, respectively. (d-f) POC-VLM on the corresponding datasets.}
    \label{fig: ConfusionMatrix}
\end{figure*}

Figure \ref{fig: ConfusionMatrix} further examines the class-wise diagnostic behaviour of POC-VLM and PerceiverIO using normalized confusion matrices. The most pronounced difference appears in the incipient cavitation class, which corresponds to the transition from non-cavitation to developed cavitation and therefore exhibits relatively ambiguous acoustic characteristics. On Cavitation-Short, PerceiverIO correctly identifies only 41.48$\%$ of incipient cavitation samples, whereas POC-VLM increases the corresponding recall to 87.50$\%$. On Cavitation-Long, the recall of the incipient cavitation class increases from 58.16$\%$ for PerceiverIO to 88.89$\%$ for POC-VLM. For Cavitation-Noise, POC-VLM correctly identifies all four cavitation states. These class-wise results indicate that the proposed framework is particularly effective for ambiguous transitional states, where acoustic patterns alone may not provide sufficient diagnostic evidence and physical operating conditions can provide complementary constraints.

\subsection{Ablation and Sensitivity Analysis}
\label{subsec: Ablation and Sensitivity Analysis}
To evaluate the contributions of the two complementary diagnostic pathways, Table \ref{tab:model_variants} reports the results obtained by removing either the vision-language pathway or the physical operating-condition prior. All experiments are repeated three times and reported as mean $\pm$ standard deviation. When the vision-language pathway is removed and diagnosis relies only on the physical operating-condition prior, the accuracy decreases from 0.9714 $\pm$ 0.0040 to 0.9104 $\pm$ 0.0018. Conversely, removing the physical operating-condition prior reduces the accuracy to 0.8859 $\pm$ 0.0049 and the F1-score from 0.9608 $\pm$ 0.0038 to 0.8384 $\pm$ 0.0051. Neither pathway alone reaches the performance of the complete model, demonstrating that acoustic semantic evidence and physical operating-condition evidence provide complementary diagnostic information. In particular, the larger degradation after removing the physical prior indicates that operating-condition information plays an important role in resolving acoustic responses that are ambiguous under different physical states.
\begin{table}[htbp]
\centering
\caption{Ablation results of different model variants of POC-VLM on Cavitation-Short. The results are averaged over three independent runs.}
\footnotesize
\label{tab:model_variants}
\setlength{\tabcolsep}{1.5mm}

\begin{tabular}{ccccc}
\toprule
Model Variant
& Accuracy
& Precision
& Recall
& F1-score \\
\midrule

w/o Vision-Language Pathway
& 0.9104 $\pm$ 0.0018
& 0.8985 $\pm$ 0.0048
& 0.8732 $\pm$ 0.0043
& 0.8800 $\pm$ 0.0050 \\

w/o Physical Operating-Condition Prior
& 0.8859 $\pm$ 0.0049
& 0.8864 $\pm$ 0.0044
& 0.8243 $\pm$ 0.0049
& 0.8384 $\pm$ 0.0051 \\

\rowcolor{gray!15}
\textbf{POC-VLM}
& \textbf{0.9714 $\pm$ 0.0040}
& \textbf{0.9694 $\pm$ 0.0058}
& \textbf{0.9549 $\pm$ 0.0017}
& \textbf{0.9608 $\pm$ 0.0038} \\
\bottomrule
\end{tabular}
\end{table}
\begin{table}[htbp]
\centering
\caption{Ablation analysis of textual semantic representation and the physical operating-condition prior on Cavitation-Short. The results are averaged over three independent runs.}
\label{tab:semantic_prior_ablation}
\footnotesize
\setlength{\tabcolsep}{1.7mm}
\begin{tabular}{cccccc}
\toprule
Physical Prior &
Textual Description &
Accuracy &
Precision &
Recall&
F1-score\\
\midrule

\ding{55} &
Static class  &
0.8316 $\pm$ 0.0031&
0.8359 $\pm$ 0.0036&
0.7248 $\pm$ 0.0032&
0.7342 $\pm$ 0.0025\\

\ding{55} &
Dynamic POC-aware  &
0.8859 $\pm$ 0.0049 &
0.8864 $\pm$ 0.0044 &
0.8243 $\pm$ 0.0049 &
0.8384 $\pm$ 0.0051 \\

\ding{51}&
Static class  &
0.9422 $\pm$ 0.0028&
0.9232 $\pm$ 0.0027&
0.9320 $\pm$ 0.0024&
0.9272 $\pm$ 0.0035\\

\rowcolor{gray!15}
\ding{51} &
Dynamic POC-aware  &
\textbf{0.9714 $\pm$ 0.0040} &
\textbf{0.9694 $\pm$ 0.0058} &
\textbf{0.9549 $\pm$ 0.0017} &
\textbf{0.9608 $\pm$ 0.0038} \\

\bottomrule
\end{tabular}
\end{table}
\begin{figure*}[htbp]
    \centering
    \includegraphics[width=\textwidth,height=75mm]{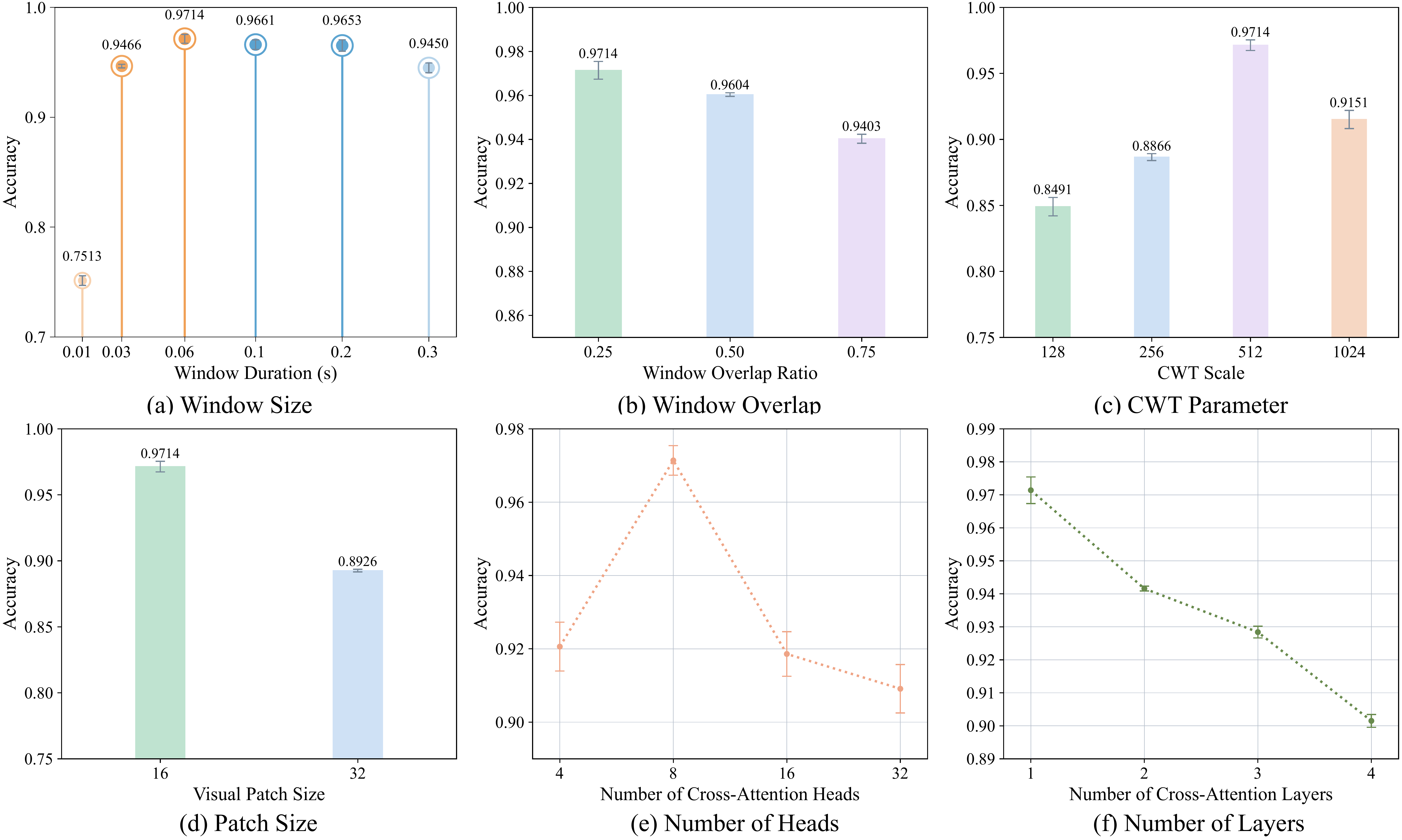}
    \caption{Ablation and sensitivity analysis of POC-VLM on Cavitation-Short. (a) Window duration, (b) window overlap ratio, (c) number of CWT scales, (d) CLIP visual patch size, (e) number of cross-attention heads and (f) number of P2Tformer layers. The results are averaged over three independent runs.}
    \label{fig: AblationAnalysis}
\end{figure*}

Table \ref{tab:semantic_prior_ablation} further examines the effect of incorporating physical operating conditions into the textual semantic representation. Without the physical operating-condition prior, replacing static class descriptions with dynamic POC-aware descriptions improves the accuracy from 0.8316 to 0.8859 and the F1-score from 0.7342 to 0.8384, indicating that sample-specific operating conditions provide more informative semantic context for interpreting acoustic time-frequency responses. This advantage remains when the physical prior is introduced, with the accuracy and F1-score further increasing from 0.9422 and 0.9272 for static class descriptions to 0.9714 and 0.9608 for dynamic POC-aware descriptions, respectively. Meanwhile, introducing the physical prior consistently improves both textual representation settings. These results demonstrate that physical operating conditions contribute at two complementary levels. Dynamic POC-aware descriptions enhance the semantic representation of candidate states, while the physical operating-condition prior provides additional decision-level evidence based on the training-condition distribution.

Figure \ref{fig: AblationAnalysis} further investigates the sensitivity of POC-VLM to signal preprocessing and model configurations. Increasing the window duration from \SI{0.01}{\s} to \SI{0.06}{\s} substantially improves the diagnostic accuracy from 0.7513 to 0.9714. Further increasing the duration does not provide additional improvement and gradually reduces performance, suggesting that a \SI{0.06}{\s} window provides an appropriate balance between capturing cavitation-related temporal characteristics and maintaining local response resolution. An overlap ratio of 0.25 achieves the highest accuracy of 0.9714, while larger overlaps lead to slight performance degradation, indicating that excessive overlap mainly introduces redundant neighbouring samples.

The CWT resolution also has a clear influence on performance. Increasing the number of scales from 128 to 512 consistently improves the results, with 512 scales achieving the highest accuracy of 0.9714. Increasing the number further to 1024 reduces the accuracy to 0.9151, suggesting that excessively fine-scale representations may introduce redundant time-frequency details. A similar trend is observed for the CLIP visual patch size, where the patch size of 16 substantially outperforms 32, indicating that finer local partitions are advantageous for capturing localized cavitation patterns. For the time-frequency-conditioned text adaptation module, eight cross-attention heads provide the best performance, while increasing the number of P2Tformer layers leads to a consistent decrease in accuracy. These results suggest that a lightweight cross-modal adaptation module is sufficient to align physical operating-condition-aware semantics with acoustic time-frequency representations, whereas excessive architectural complexity does not provide additional benefit.

\subsection{Analysis of Physical Operating Conditions}
\label{subsec: Analysis of Physical Operating Conditions}
\begin{figure*}[htbp]
    \centering
    \includegraphics[width=\textwidth,height=40mm]{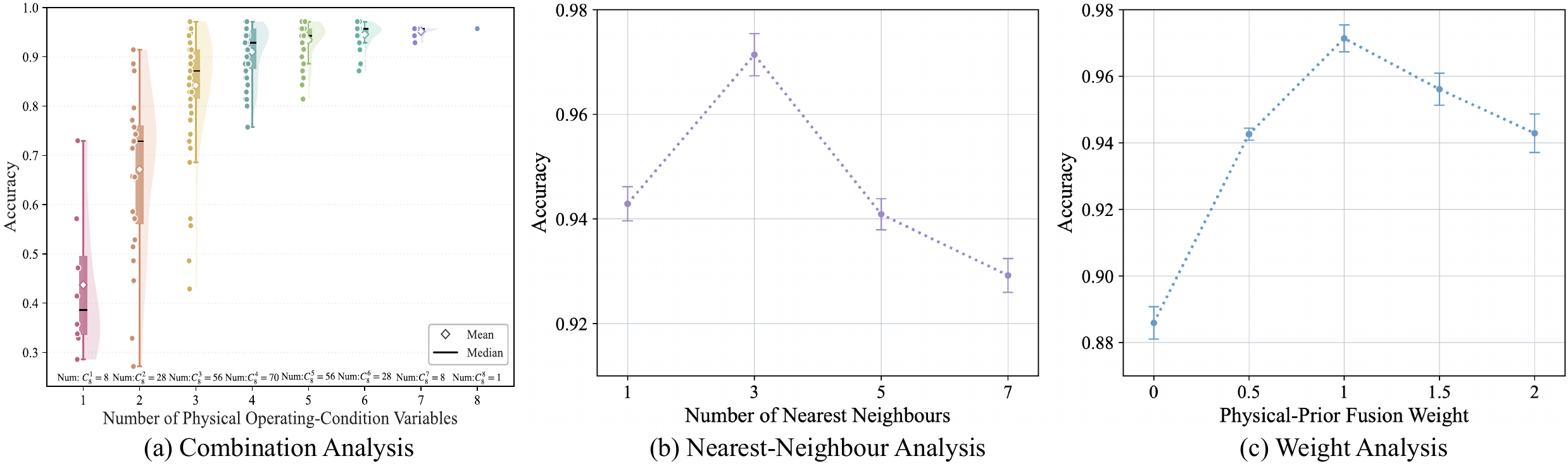}
    \caption{Analysis of physical operating conditions for POC-VLM on Cavitation-Short. (a) Combination analysis of the 255 non-empty combinations of eight physical operating-condition variables, (b) nearest-neighbour analysis and (c) physical-prior fusion weight analysis. The results are averaged over three independent runs.}
    \label{fig: PhysicalConditionAnalysis}
\end{figure*}

To further investigate how physical operating conditions contribute to POC-VLM, Figure \ref{fig: PhysicalConditionAnalysis}(a) evaluates all 255 non-empty combinations formed by the eight available physical variables. The results show a clear relationship between the amount of physical information and diagnostic performance.When only a single physical variable is used, the performance varies considerably across different choices, indicating that no individual variable can fully characterize the operating state associated with cavitation development. As additional variables are introduced, both the overall accuracy and the lower bound of the performance distribution increase substantially. The improvement becomes progressively smaller after approximately three variables, indicating that a compact subset of operating variables can already provide sufficient physical constraints for the current diagnostic task.

Table \ref{tab: BestPhysicalCombination} reports the best-performing variable combination for each number of physical operating-condition variables. Among the single-variable settings, the pressure recovery factor $X_f$ provides the highest accuracy of 0.7297. Combining $X_f$ with the valve flow coefficient $K_v$ increases the accuracy to 0.9143. With three variables, the combination of $X_f$, valve stroke $h$ and upstream pressure $p_u$ reaches an accuracy of 0.9714, which is identical to the result obtained with larger variable subsets. Notably, high diagnostic performance can also be achieved when $X_f$ is excluded and replaced by other physical operating-condition variables, indicating that the model does not rely exclusively on this derived quantity. This is reasonable because $X_f$ itself is calculated from the upstream pressure, downstream pressure and vapour pressure, and therefore partially overlaps with the information contained in the original pressure measurements. These results suggest that physical operating-condition information contains substantial redundancy and that a small number of complementary variables can already characterize the operating state effectively.
\begin{table}[htbp]
\caption{Best-performing physical operating-condition variable combinations for different numbers of variables on Cavitation-Short.}
\label{tab: BestPhysicalCombination}
\footnotesize
\centering
\setlength{\tabcolsep}{2.5mm}
\begin{tabular}{ccccccccccccc}
\toprule
\multirow{2}{*}{Number} 
& \multicolumn{8}{c}{Physical Operating Conditions} 
& \multicolumn{4}{c}{Evaluation Metrics}       \\ 
\cmidrule{2-13} 

& ${X_f}$     & $K_v$    & $h$    & $p_u$    & $p_d$   & $Q$   & $p_v$   & $T$   
& Accuracy    & Precision     & Recall       & F1-score     \\ 
\midrule
1       
& \ding{51} &  &  & &  &  &  &    
& 0.7297    & 0.6341     & 0.6312     & 0.6262        \\

2       
& \ding{51} & \ding{51} &  &  &  & &  &    
& 0.9143    & 0.9248     & 0.8701     & 0.8856        \\

3
& \ding{51} &  & \ding{51} & \ding{51} &  &  &  &   
& \textbf{0.9714}    & \textbf{0.9694}     & \textbf{0.9549}    & \textbf{0.9608}       \\

\rowcolor{gray!15}
4       
&  &  & \ding{51} & \ding{51} & \ding{51} & \ding{51} &  &   
& \textbf{0.9714}    & \textbf{0.9694}     & \textbf{0.9549}    & \textbf{0.9608}       \\

\rowcolor{gray!15}
5
& & \ding{51} & \ding{51} & \ding{51} & \ding{51} & \ding{51} &  &      
& \textbf{0.9714}    & \textbf{0.9694}     & \textbf{0.9549}    & \textbf{0.9608}       \\

6      
& \ding{51} & \ding{51} & \ding{51} & \ding{51} & \ding{51} & \ding{51} & & 
& \textbf{0.9714}    & \textbf{0.9694}     & \textbf{0.9549}    & \textbf{0.9608}       \\

7       
& \ding{51} &  & \ding{51} & \ding{51} & \ding{51} & \ding{51} & \ding{51} & \ding{51}      
& \textbf{0.9714}    & \textbf{0.9694}     & \textbf{0.9549}    & \textbf{0.9608}       \\

8     
& \ding{51} & \ding{51} & \ding{51} & \ding{51} & \ding{51} & \ding{51} & \ding{51} & \ding{51}      
& \textbf{0.9714}    & \textbf{0.9694}     & \textbf{0.9549}    & \textbf{0.9608}       \\
\bottomrule
\end{tabular}
\end{table}

Figure \ref{fig: PhysicalConditionAnalysis}(b) evaluates the influence of the number of nearest neighbours used to estimate the physical operating-condition prior. The best performance is obtained at $k=3$. Using only one neighbour makes the prior more sensitive to an individual operating-condition prototype, whereas increasing $k$ beyond three gradually reduces performance because less similar operating conditions are incorporated into the voting process. This result supports the use of a small local neighbourhood to characterize the relationship between operating conditions and cavitation states.

The contribution of the physical operating-condition prior is further examined by varying the fusion coefficient $\lambda_{\mathrm{POC}}$, as shown in Figure \ref{fig: PhysicalConditionAnalysis}(c). When $\lambda_{\mathrm{POC}}=0$, the physical prior is removed and the model reduces to the vision-language pathway, resulting in an accuracy of 0.8859. Increasing $\lambda_{\mathrm{POC}}$ progressively improves performance and reaches the maximum at $\lambda_{\mathrm{POC}}=1$. Further increasing the weight causes a moderate decline. This behaviour indicates that the physical prior provides important complementary constraints, but excessively emphasizing it can suppress the semantic evidence extracted from the acoustic response. The best performance at $\lambda_{\mathrm{POC}}=1$ reflects an appropriate balance between semantic compatibility and physical plausibility.

\subsection{Generalization and Robustness Analysis}
\label{subsec: Generalization and Robustness Analysis}
\begin{figure*}[htbp]
    \centering
    \includegraphics[width=\textwidth,height=40mm]{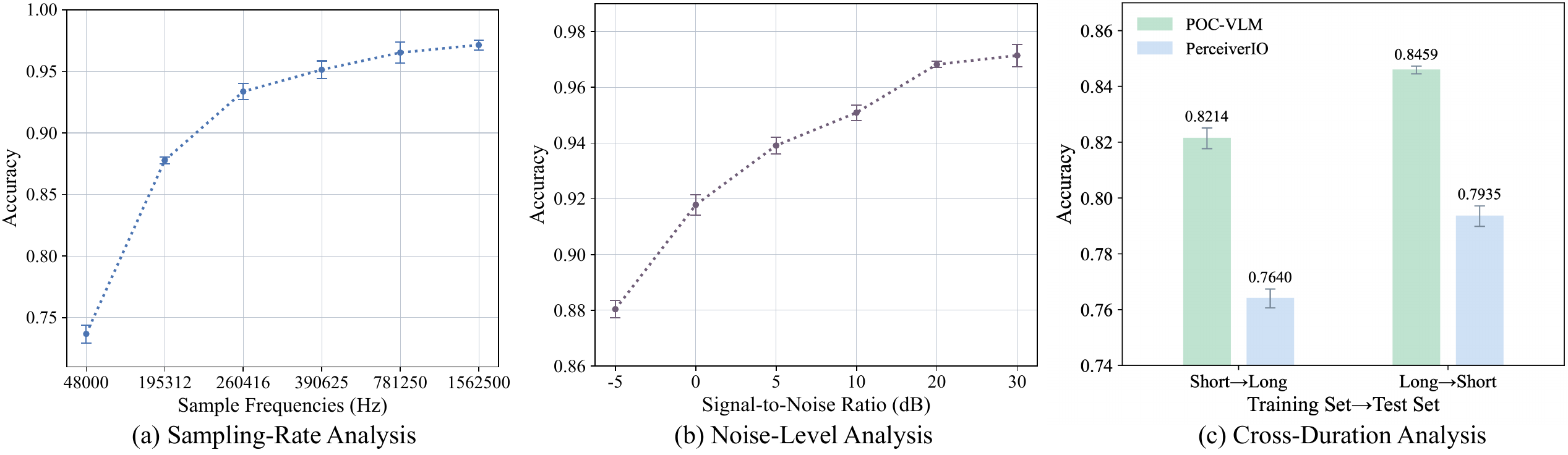}
    \caption{Generalization and robustness analysis of POC-VLM. (a) Sampling-rate analysis and (b) noise-level analysis on Cavitation-Short, and (c) cross-duration analysis between Cavitation-Short and Cavitation-Long. The results are averaged over three independent runs.}
    \label{fig: RobustnessAnalysis}
\end{figure*}

Figure \ref{fig: RobustnessAnalysis} evaluates the generalization and robustness of POC-VLM under variations in sampling frequency, signal-to-noise ratio and cross-duration. In practical applications, the ability to recognize acoustic signals acquired by low-cost and low-sampling-rate sensors is important for deployment-oriented diagnosis. To evaluate this capability, POC-VLM is tested at the original sampling frequency of \SI{1562.5}{\kHz}, as well as at one-half, one-quarter, one-sixth and one-eighth of the original sampling frequency, corresponding to 781.25, 390.625, 260.416 and 195.312 \si{\kHz}, respectively. We further consider \SI{48}{\kHz}, approximately one thirty-second of the original sampling frequency, which represents a typical upper sampling rate supported by consumer-grade mobile devices. From Figure \ref{fig: RobustnessAnalysis}(a), the sampling-rate analysis shows that diagnostic performance improves consistently with increasing sampling frequency. At the lowest sampling frequency of \SI{48}{\kHz}, the accuracy decreases to 0.7368, indicating that substantial high-frequency cavitation information is lost. Increasing the sampling frequency to approximately \SI{260.416}{\kHz} already recovers an accuracy above 0.9336, while further increases provide progressively smaller gains and eventually approach the original performance. These results indicate that high-frequency acoustic components contribute substantially to cavitation intensity discrimination, while POC-VLM can still maintain useful diagnostic capability under considerably reduced sampling rates.

As shown in Figure \ref{fig: RobustnessAnalysis}(b), the noise-level analysis evaluates the robustness of POC-VLM under additive noise with signal-to-noise ratios ranging from -5 to 30 dB. The diagnostic accuracy decreases gradually as the noise level increases, from 0.9714 at 30 dB to 0.9682, 0.9509, 0.9391, 0.9178 and 0.8804 at 20, 10, 5, 0 and -5 dB, respectively. Notably, the model maintains an accuracy above 0.95 at 10 dB and still achieves 0.8804 under the severe -5 dB condition. The relatively smooth degradation across different noise levels indicates that POC-VLM retains stable diagnostic capability when the acoustic observations are progressively corrupted. This robustness can be attributed to the complementary use of acoustic semantic evidence and physical operating-condition information, which reduces the dependence of the final decision on acoustic observations alone.

Figure \ref{fig: RobustnessAnalysis}(c) further evaluates cross-duration generalization by exchanging the training and testing datasets between Cavitation-Short and Cavitation-Long. When trained on Cavitation-Short and evaluated on Cavitation-Long, POC-VLM achieves an accuracy of 0.8214, compared with 0.7640 for PerceiverIO. In the reverse Cavitation-Long to Cavitation-Short setting, POC-VLM achieves 0.8459, outperforming PerceiverIO at 0.7935. Although the performance of both methods decreases relative to the corresponding within-dataset evaluation, POC-VLM consistently maintains higher accuracy under both transfer directions. These results indicate that changes in observation duration introduce a clear distribution shift in the acoustic representations, while the incorporation of physical operating-condition constraints improves the generalization of POC-VLM across different observation durations.
\section{Conclusions}
\label{sec: conclusions}
This work presents a physical operating-condition-guided vision-language matching for cavitation diagnosis in industrial pipeline systems (POC-VLM). POC-VLM constructs physical operating condition-aware semantic descriptions, adapts them according to acoustic time-frequency responses and performs global-local vision-language matching. In parallel, a KNN-based non-parametric physical prior is introduced to complement semantic evidence with physical plausibility, enabling joint reasoning from acoustic responses and operating conditions. Extensive experiments on three real-world cavitation datasets demonstrate that POC-VLM consistently outperforms representative modern and domain-specific diagnostic methods. Further analyses confirm the complementary roles of the vision-language pathway and physical operating-condition prior, while showing that compact physical-variable subsets can already provide effective operating-condition constraints. The method also exhibits favourable robustness under reduced sampling frequencies, additive noise and cross-duration transfer.

Future work will focus on improving the generalization of POC-VLM under larger variations in sensing and operating conditions. More flexible physical-semantic representation strategies and adaptive physical-prior estimation will be investigated to reduce dependence on predefined operating variables and fixed neighbourhood configurations. In addition, extending the framework to different industrial systems, sensing modalities and condition-dependent diagnostic tasks will be explored to further evaluate its transferability and applicability in broader industrial environments.

\section*{CRediT authorship contribution statement}
\label{sec: authorship contribution statement}
\textbf{Yu Sha}: Conceptualization, Methodology, Programming, Calculation, Data curation, Writing - Original Draft. \textbf{Junqi Tao}: Programming, Formal analysis, Visualization. \textbf{Deyu Wang}: Programming, Formal analysis, Visualization. \textbf{Horst Stoecker}: Writing - Review \& Editing, Supervision, Funding acquisition. \textbf{Domagoj Vnucec}: Resources, Supervision, Funding acquisition. \textbf{Andreas Widl}: Resources, Supervision, Funding acquisition. \textbf{Kai Zhou}: Conceptualization, Writing - Review \& Editing, Supervision, Project administration, Funding acquisition.

\section*{Declaration of competing interest}
The authors declare that they have no known competing financial interests or personal relationships that could have appeared to influence the work reported in this paper.

\section*{Declaration of Generative AI and AI-assisted technologies in the writing process}
\label{sec: Declaration of Generative AI}
\textbf{Author's contribution:} During the preparation of this work, the authors used ChatGPT (GPT 6) to improve language fluency and enhance readability. After using this tool, the authors reviewed and edited the content as necessary and take full responsibility for the content of published article.

\textbf{Statement:} The tool was used solely for language enhancement. No AI was involved in data generation, image generation, analysis or interpretation of results.

% Acknowledgements
\section*{Acknowledgements}
\begingroup
\setlength{\emergencystretch}{3em}
This research is supported by the China Postdoctoral Science Foundation under Grant No.\ 2025M781519 (Y.S.), the National Natural Science Foundation of China under Grant No.\ 92570117, the Ministry of Science and Technology of China under Grant No.\ 2024YFA1611004, the Shenzhen Peacock Fund under Grant Nos.\ 2023TC0007 and 2023TC0179, the CUHK-Shenzhen University Development Fund under Grant Nos.\ UDF01003041 and UDF03003041 (Y.S., K.Z.), the AI grant at FIAS through SAMSON AG (K.Z.), the BMBF-funded KISS consortium (05D23RI1) in the ErUM-Data action plan (K.Z.), SAMSON AG (D.V., A.W.), the Walter Greiner Gesellschaft zur F\"orderung der physikalischen Grundlagen\-forschung e.V. through the Judah M. Eisenberg Laureatus Chair at Goethe Universit\"at Frankfurt am Main (H.S.), and the NVIDIA GPU grant through NVIDIA Corporation (K.Z.).
\par
\endgroup

\section*{Code availability}
\label{sec: Code availability}
The source code for POC-VLM is publicly available at
\textcolor{orange}{\url{https://github.com/CavitationDetection/POC-VLM}}.

%% If you have bib database file and want bibtex to generate the
%% bibitems, please use
%%
% The bibliography is disabled while the manuscript contains no citations.
% Uncomment both lines after adding at least one \cite{...} command.
\bibliographystyle{elsarticle-num-names}
\bibliography{refss}

% \clearpage
% \input{section/appendix.tex}

\end{document}